# Two-Phase Object-Based Deep Learning for Multi-temporal SAR image change Detection

Xinzheng Zhang[1,2,*], Guo Liu[1], Ce Zhang[3,4*], Peter M Atkinson[3], Xiaoheng Tan[1,2], Xin Jian[1,2], Xichuan Zhou[1], Yongming Li[1]

[1] College of Microelectronics and Communication Engineering, Chongqing University, Chongqing, China;

[2] Chongqing Key Laboratory of Space Information Network and Intelligent Information Fusion, Chongqing, China;

[3] Lancaster Environment Centre, Lancaster University, Lancaster LA1 4YQ, United Kingdom;

[4] UK Centre for Ecology & Hydrology, Library Avenue, Lancaster LA1 4AP, United Kingdom;

* Correspondence e-mail: zhangxinzheng@cqu.edu.cn; c.zhang9@lancaster.ac.uk

**Abstract:** Change detection is one of the fundamental applications of synthetic aperture radar (SAR) images. However, speckle noise presented in SAR images has a negative effect on change detection, leading to frequent false alarms in the mapping products. In this research, a novel two-phase object-based deep learning approach is proposed for multi-temporal SAR image change detection. Compared with traditional methods, the proposed approach brings two main innovations. One is to classify all pixels into three categories rather than two categories: unchanged pixels, changed pixels caused by strong speckle (false changes), and changed pixels formed by real terrain variation (real changes). The other is to group neighboring pixels into segmented into superpixel objects (from pixels) such as to exploit local spatial context. Two phases are designed in the methodology: 1) Generate objects based on the simple linear iterative clustering (SLIC) algorithm, and discriminate these objects into changed and unchanged classes using fuzzy c-means (FCM) clustering and a deep PCANet. The prediction of this Phase is the set of changed and unchanged superpixels. 2) Deep learning on the pixel sets over the *changed* superpixels only, obtained in the first phase, to discriminate real changes from false changes. SLIC is employed again to achieve new superpixels in the second phase. Low rank and sparse decomposition are applied to these new superpixels to suppress speckle noise significantly. A further clustering step is applied to these new superpixels via FCM. A new PCANet is then trained to classify two kinds of changed superpixels to achieve the final change maps. Numerical experiments demonstrate that, compared with benchmark methods, the proposed approach can distinguish real changes from false changes effectively with significantly reduced false alarm rates, and achieve up to 99.71% change detection accuracy using multi-temporal SAR imagery.

## 1. Introduction

With its cloud penetrating capability, synthetic aperture radar (SAR) images have drawn a large amount of attention, for example, in environmental surveillance, urban planning and military applications over the past decades. Using SAR images for change detection often involves two images acquired over the same area at different times, utilising the information in the differences between them.

Depending on the availability of a difference image (DI), change detection approaches can be divided into two categories. One is post-classification comparison which is undertaken to identify changed and unchanged regions directly from two images that were classified independently before the analysis. In this approach, the change detection result is not influenced by radiation normalization and geometric correction. However, the accuracy of the change detection relies on the quality of the classification results, with errors propagating to the outcome. The other approach is post-comparison analysis, in which change detection is achieved by generating a DI from two multi-temporal images, and obtaining the final change map from it. The classification errors in this case do not accumulate, but the way that the DI is generted may influence the validity of the change detection results [1].

From a machine learning perspective, change detection can also be categorized into supervised and unsupervised approaches, depending on whether labeled data are used or not [2-3]. For supervised methods, features extracted from labeled data are fed into a subsequent classifier. This strategy requires a significant number of ground reference data to train the algorithm, and the labelling process can be extremely labor-intensive and time-consuming [4]. In [5], a context-sensitive similarity measure is presented based on supervised classification to amplify the dissimilarity between changed and unchanged pixels. Unsupervised methods for change detection can be viewed as a clustering approach which divides the data into changed and unchanged classes [6-7]. In [8], the DI is cast into an eigenvector space and $k$-means clustering is used to partition the space into two clusters. In [9], a modified Markov Random Field (MRF) energy function is employed to update iteratively the membership association of fuzzy $c$-means (FCM), to cluster the DI into two classes. In [10] a novel method based on spatial fuzzy clustering was used to add spatial information to enhance change detection performance.

Recently, deep learning has gained widespread attention in the field of computer vision and pattern recognition, and demonstrated state-of-the-art prediction accuracy in various challenging tasks, such as target detection, image classification, etc.. The major benefit of deep learning is that it can extract abstract and high-level representations that are hard to hand-code through feature engineering [11,12]. Besides, deep networks are often pre-trained using a large-scale dataset (e.g. ImageNet), and fine-tuned to other domains including remote sensing. Convolutional neural networks (CNNs) are considered as the pioneer of deep learning methods which mimic the receptive fields of the human brain neural cortex, with less redundancy and complexity through the weight-sharing architecture [12,13]. Some well-developed CNN models, such as AlexNet [12], VGG [14] and ResNet [15], have been adopted quickly in the remote sensing community to solve real-world challenges (e.g., land cover and land use classification).

Given the advantages of deep learning, some pioneering methods have been proposed for multi-temporal SAR image change detection. In [1], a stack of restricted Boltzmann machine (RBM) networks was used to learn efficiently the relationship between two multi-temporal SAR images for change detection. A dual-channel CNN structure was used to extract features of two SAR images for change detection [16]. [17] presents a local restricted CNN for SAR image change detection, which is formed by imposing a spatial constraint on the output layer of the CNN, such as to learn from several layered difference images. In [18], a stacked contractive autoencoder (sCAE) using a contractive penalty was proposed to promote local invariance and robustness, such that robust features can be extracted from superpixels of SAR images for change detection. In [19], a deep learning-based weakly supervised framework was developed for urban change detection using multi-temporal polarimetric SAR data. In [20], a transferred multi-level fusion network (MLFN) was trained using a large dataset and fine-tuned to extract features from SAR image patches for sea ice change detection. PCANet is an alternative deep learning model suitable for SAR image change detection [22,23,24]. In PCANet, the cascaded PCA filters and binary quantization (hashing) are used as a data-adapting convolution filter bank in each stage and in the nonlinearity layer [21]. During the PCANet training process, there is no requirement for regularized parameters and numerical optimization solvers, which promotes the efficiency and accuracy of the network. In [22], PCANet was shown to be accurate, with great potential for SAR image change detection. In [23], context-aware saliency detection was employed to obtain training samples for PCANet in SAR image change detection, which reduces the number of training samples required while maintaining the reliability of the training sample sets, leading to less training time and computational efficiency. In [24], a morphologically supervised PCANet was designed to overcome the class imbalance problem in SAR image change detection (changed pixels are far less common than unchanged pixels).

Although the above-mentioned deep learning methods exhibit excellent performance in SAR image change detection, there are still some shortcomings. First of all, all the above methods are actually binary classification algorithms, which separate pixels of the changed class (CC) from pixels of the unchanged class (UC). In reality, variation in the pixel values caused by strong speckle noise may lead to allocation to the changed class, potentially producing a large number of false alarms. There are actually two kinds of changed pixels: one is produced by real terrain object changes (i.e. real changed class, RCC), and the other caused by strong speckle noise (i.e. false changed class, FCC). Even if deep learning models have powerful classification capabilities, there will still be several false alarms due to strong speckle noise. Secondly, in current deep learning-based SAR image change detection, high quality training samples are required to train the networks. Those training samples are commonly taken as rectangular patches centering around the pixels that are of interest. However, this operation often introduces artefacts on the border of these rectangular patches, which produces uncertainty in the classification maps. For example, unchanged pixels and changed pixels could potentially exist in one image patch simultaneously. Heterogeneous pixels can also be found in one rectangular patch, which will increase the difficulty of distinguishing between CC and UC classes.

In this research, a new framework of two-phase object-based deep learning (TPOBDL) is proposed for SAR image change detection. Object-based deep learning has been shown to be suitable for remote sensing applications [25]. Thus, in TPOBDL, change detection is implemented in an object-based rather than pixel-wise fashion. Superpixel generation is applied to SAR images to acquire image objects (also called superpixels in computer science, and here) using a simple linear iterative clustering (SLIC) algorithm [26]. In fact, all processing steps in TPOBDL are based on image superpixels. Since a superpixel is a local set of homogeneous pixels, superpixels can reflect the local spatial context [27,28,29]. Therefore, this approach can overcome the problems caused by operations involving rectangular patches, such as introducing artefacts and uncertainty in the classification. The proposed approach involves two phases to differentiate RCC and FCC objects in an automated approach. Our two-phase deep learning strategy is, thus: Phase 1 deep learning to classify the objects of CC and those of UC, and Phase 2 deep learning to classify objects of CC into RCC and FCC objects. This two-phase framework reduces the classification difficulty faced by deep learning models at each phase, and is conducive to increasing the overall accuracy of change detection.

Our major contributions are as follows:

1) Change detection through an object-based rather than pixel-wise approach. Superpixel generation is applied to SAR images to obtain objects via SLIC, such that the local spatial context is captured.

2) A two-phase approach is designed for multi-temporal SAR image change detection. Deep learning methods are developed to identify objects of FCC and RCC by combining low rank and sparse decomposition (LRSD) with reduced false alarms.

The remainder of this paper is organized as follows. In Section 2, the proposed approach is described in detail. Section 3 presents the experimental datasets and results. Discussion on the experiment results and the proposed approach are shown in Section 4. Finally, conclusions are drawn in Section 5.

## 2. Methodology

### 2.1. Problem Statement and Overview of the Proposed Method

Consider two SAR images taken from the same location, but at different times $I_1$ and $I_2$, both of size $M \times N$. Change detection is required to generate a binary change map labeling changed pixels and unchanged pixels between $I_1$ and $I_2$. Figure 1 shows the scheme of TPOBDL, which consists mainly of two phases of deep learning, described in detail as follows.

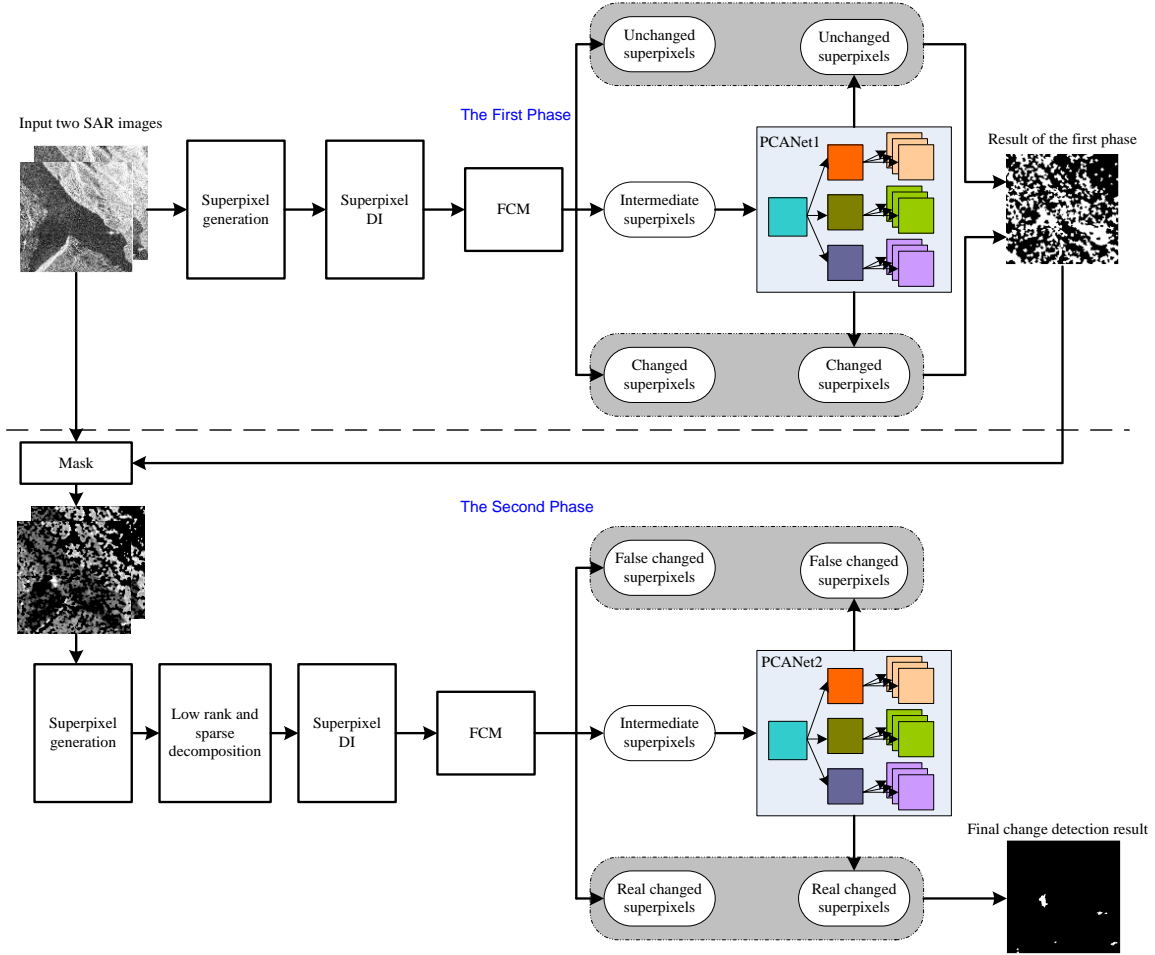

**Figure 1.** The scheme of the proposed approach.

### 2.2. First Phase Deep Learning

2.2.1. Superpixel Generation of Multi-Temporal SAR Images

In existing deep learning-based SAR image change detection methods, the patches for the training and testing of deep neural networks are generated mainly in the shape of rectangles, which is convenient [24]. However, the operation of taking rectangular patches has significant disadvantages for SAR image change detection. Firstly, when the current pixel is near the boundary between changed and unchanged regions, the patch generated will contain both changed and unchanged pixels, which may introduce uncertainty to the deep neural network and impair the learning process [25]. Secondly, rectangular patch generation ignores the local spatial context, which is conducive to the change detection. Instead of taking a rectangular patch, in this paper, patches come from superpixels, where all pixels are homogeneous. This reduces the likelihood that heterogeneous pixels, or even changed and unchanged pixels appear in one patch simultaneously. Patches that are superpixels, compared with traditional rectangular patches, provide more valid information to the deep learning model. In fact, deep learning based on superpixels is an object-based approach, which have more advantages.

In this research, we use SLIC to apply superpixel generation to two multitemporal SAR images $I_1$ and $I_2$. SLIC is chosen for its simplicity, flexibility in compactness, memory efficiency and high accuracy, as applied to SAR image processing [30,31]. First, superpixels of $I_1$ are obtained by SLIC. Then the superpixel pattern from $I_1$ is copied to $I_2$, as shown in Fig 2. Pattern copying ensures that the corresponding two superpixels of $I_1$ and $I_2$ represent the same local region.

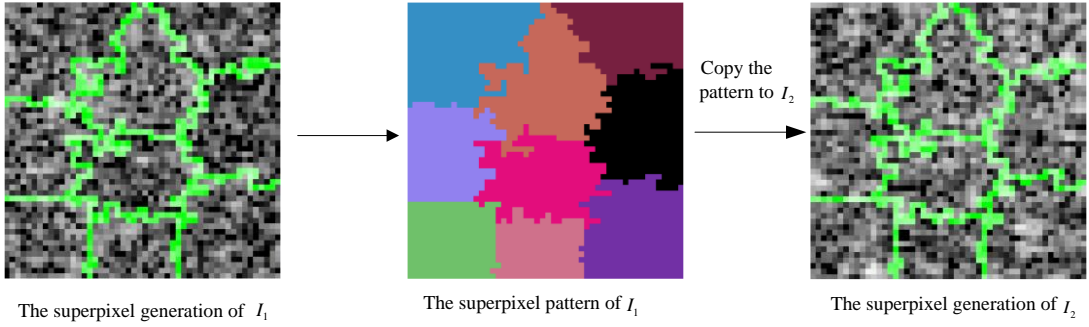

The superpixel generation of $I_1$    The superpixel pattern of $I_1$    The superpixel generation of $I_2$

**Figure 2.** Illustration of copying superpixel pattern from $I_1$ to $I_2$.

The principles of SLIC are briefly described as follows. Firstly, the number of superpixels is set as $v$, which means $I_1$ is portioned into $v$ pixel-blocks at the beginning. The center of each pixel-block is called a seed. The distance (step length) between two seeds is defined as $\Omega = \sqrt{M \times N / v}$. To avoid seeds falling on the contour boundary with a larger gradient, the seeds are redefined where the gradient is the smallest in the neighborhood. Then searching in the neighborhood of each seed, the distance between a pixel in the neighborhood and the seed, including distance in feature (colour) space $d_c$ and in geographical space $d_s$, is gained by

$$d_c = \sqrt{(l_j - l_i)^2 + (a_j - a_i)^2 + (b_j - b_i)^2} \quad (1)$$

$$d_s = \sqrt{(x_j - x_i)^2 + (y_j - y_i)^2} \quad (2)$$

$$D = \sqrt{(d_c/10)^2 + (d_s/\Omega)^2} \quad (3)$$

where $d_c$ means feature (color) distance, $d_s$ means spatial distance, and $D$ is the distance metric. $l_i$, $a_i$ and $b_i$ represent the three principal elements in the Lab module of the seed, and $x_i$, $y_i$ represents the coordinate of the seed. $l_j$, $a_j$, $b_j$, $x_j$ and $y_j$ are corresponding parameters of the pixel in the neighborhood. In this manner, a pixel will be searched many times with different seeds. The seed with the smallest $D$ is taken as the clustering center of this pixel. Then the seeds are updated. Generally, 10 iterations is enough to obtain satisfactory superpixels.

Superpixels possess a range of geometries and sizes (i.e., numbers of pixels). In contrast, the inputs of the deep neural network are required to be uniform rectangles with the same numbers of pixels. Thus, the superpixels need to be reshaped into rectangles before being fed into the network. Assume that the input patches are of size $k \times k$. Then, each reshaped superpixel should also have $k^2$ pixels. If a superpixel contains $p$ pixels, there are two ways to reshape the superpixel. One is $p \leq k^2$. For this case, assume that a superpixel represented as $S_{n,i}^m$ (where $m$ represents the phase it is in, in this stage $m=1$, $n$ represents the image it comes from, $n=1,2$, $i$ is an index of the superpixels, $i=1,2,\cdots,v$) is reshaped to a vector $V_{n,i}^m$ having $k^2$ pixels. The first $p$ pixels of $V_{n,i}^m$ is filled by pixels of $S_{n,i}^m$, and the other $k^2 - p$ pixels are chosen randomly from $S_{n,i}^m$. The other one is $p > k^2$. For this case, we reshape the superpixel $S_{n,i}^m$ into $q+1$ vectors $V_{n,i,1}^m$, $V_{n,i,2}^m$, ..., $V_{n,i,q}^m$, each of which has $k^2$ pixels, and an extra vector with $p - qk^2$ pixels. This extra vector is filled with a vector $V_{b,i,(q+1)}^a$ of $k^2$ pixels under the condition $p \leq k^2$. For a unified description, $V_{n,i}^m$ of case $p \leq k^2$ is redefined as $V_{n,i,1}^m$.

2.2.2. Superpixel DI Generation and FCM

The reshaped superpixel vectors $V_{1,i,h}^1$ and $V_{2,i,h}^1$ ($h=1,2,\ldots,q,q+1$) from $S_{1,i}^1$ and $S_{2,i}^1$ of $I_1$ and $I_2$ are fed into the superpixel DI (SPDI) operator $F_{i,h}^1 = |V_{1,i,h}^1 - V_{2,i,h}^1|$. All $F_{i,h}^1$ form a SPDI. The reason for generating the superpixel difference map is to help the FCM algorithm to cluster satsifactorily in the next step. Then all the $F_{i,h}^1$ are clustered into three classes by FCM: changed class (CC) $\omega_c^1$, unchanged class (UC) $\omega_u^1$ and intermediate class $\omega_m^1$. Details of FCM can be found in [32]. $F_{i,h}^1$ belonging to $\omega_c^1$ or $\omega_u^1$ means that superpixel $S_{1,i}^1$ and $S_{2,i}^1$ corresponding to $V_{1,i,h}^1$ and $V_{2,i,h}^1$ have a high probability to be changed or unchanged, respectively. The pair of superpixels $S_{1,i}^1$ and $S_{2,i}^1$ with the case $p \leq k^2$ can easily be inferred to be one of three classes, because each pair of them only has one set of $V_{1,i,h}^1$ and $V_{2,i,h}^1$ which forms one $F_{i,h}^1$. However, for superpixels $S_{1,j}^1$ and $S_{2,j}^1$ with $p > k^2$, each pair has $q+1$ sets of $V_{1,i,h}^1$ and $V_{2,i,h}^1$, which leads to $q+1$ $F_{i,h}^1$. Thus, a voting mechanism is employed to determine their classes. Specifically, for the $+1$ $F_{i,h}^1$, those clustered into $\omega_c^1$ are weighted by 1, those clustered into $\omega_u^1$ are weighted by 0 and those clustered into $\omega_m^1$ are weighted by 0.5. Then, all $q+1$ weights are summed to be $\Lambda$, and the class of superpixel pair $S_{1,j}^1$ and $S_{2,j}^1$ with $p > k^2$ is determined as follows:

$$\text{class of superpixel pair } S_{1,j}^1 \text{ and } S_{2,j}^1 = \begin{cases} \omega_c^1, & \Lambda/(q+1) \geq 0.8 \\ \omega_m^1, & 0.8 > \Lambda/(q+1) \geq 0.5 \\ \omega_u^1, & \Lambda/(q+1) < 0.5 \end{cases} \quad (4)$$

The $V_{b,i,h}^1$ determined as CC and UC are reshaped to patches, which will be fed into the deep learning model as training samples. Those $V_{b,i,h}^1$ belonging to the intermediate class will be classified to CC or UC by the trained deep neural network.

2.2.3 Training PCANet1

As a type of deep learning model, PCANet is easy to train and can be adapted to other tasks. For SAR image change detection, PCANet has been shown to learn non-linear relations from multi-temporal SAR images, which is an advantage compared to other deep neural networks [22]. It has already been employed in SAR image change detection [22,23,24]. Considering these superiorities of PCANet in SAR image change detection tasks, we use PCANet here to further classify those superpixel pairs identified to the intermediate class in the previous phase. Since PCANet is used in the second phase, the network in the first phase is called PCANet1.

First, the $V_{b,i,h}^1$ of CC and UC are used as samples to train PCANet1. $V_{1,i,h}^1$ and $V_{2,i,h}^1$ are reshaped and combined to form the patches $R_{i,h}$ to be fed into the network (Fig. 3). If $I_1$ is segmented into $v$ superpixels and the $i$-th superpixel is reorganized as $\gamma_i$ vectors. Then the number of $R_{i,h}$ of size $2k \times k$ is $\Gamma = \sum_{i=1}^{v} \gamma_i$.

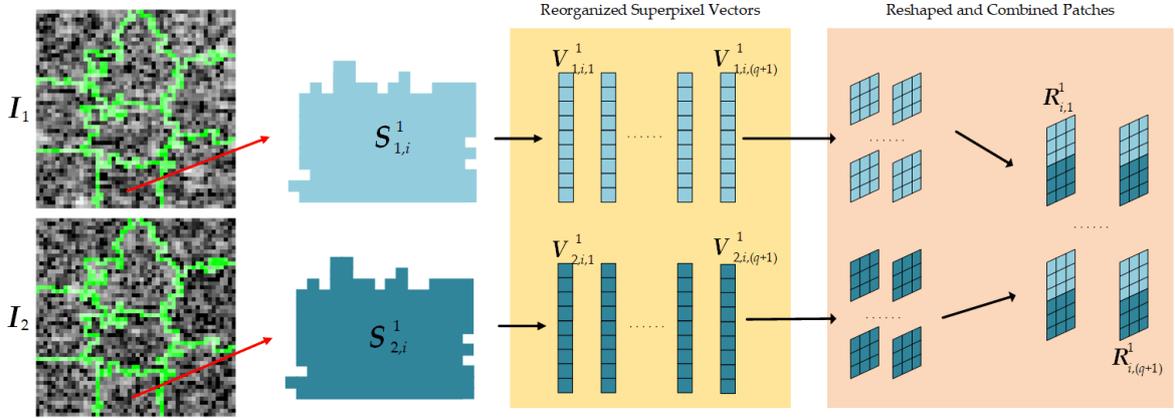

**Figure 3.** Patch generation in stage 1.

The structure of PCANet1 is shown in Fig. 4, consisting of two PCA filters convolution layers, a Hashing and histogram generation layer. After patch generation, all $R_{i,h}$ have their means removed, are vectorized and combined as a matrix $Y$.

$$Y = \left[ y_{1,1}, \ldots, y_{1,\gamma_1}, y_{2,1}, \ldots, y_{2,\gamma_2}, \ldots, y_{v,1}, \ldots, y_{v,\gamma_v} \right] \quad (5)$$

where $y_{i,h}$ denotes mean-removed and vectorized $R_{i,h}$.

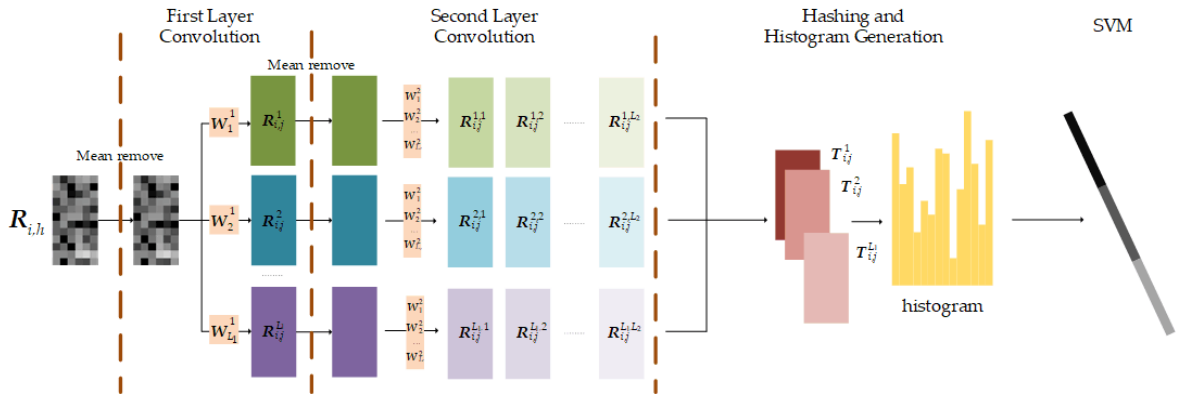

**Figure 4.** the structure of PCANet.

Next, we choose $L_1$ principal eigenvectors of $YY^T$ (T denotes the matrix transposition) as the PCA filters $W_l^1$ of the first layer, that is

$$W_l^1 = \text{mat}\left(ql(YY^T)\right) \in \Re^{2k^2 \times 2k^2}, \quad l = 1, 2, \ldots, L_1 \tag{6}$$

where $ql(YY^T)$ means $l$-th principal eigenvector and $\text{mat}(x)$ can map a vector $x \in \Re^{4k^4}$ into a matrix $W \in \Re^{2k^2 \times 2k^2}$. So, the output of the first layer is

$$R_{i,h}^l = R_{i,h} * W_l^1 \tag{7}$$

where the $*$ operator means 2-D convolution. $R_{i,h}^l$ forms the input of the second layer.

In the second layer, all $R_{i,h}^l$ have their means removed and are vectorized to be $z_{i,h}^l$, which is combined to be a matrix $Z^l = \left[z_{1,1}^l, \ldots, z_{1,\gamma_1}^l, z_{2,1}^l, \ldots, z_{2,\gamma_2}^l, \ldots, z_{v,1}^l, \ldots, z_{v,\gamma_v}^l\right]$. Then, all $Z^l$ are combined as:

$$Z = \left[Z^1, Z^2 \ldots, Z^{L_1}\right] \tag{8}$$

The following step is similar to that for the first layer. We choose $L_2$ principal eigenvectors of $ZZ^T$ as the PCA filters $W_l^2$ of the first layer, that is:

$$W_p^2 = \text{mat}\left(ql(ZZ^T)\right) \in \Re^{2k^2 \times 2k^2}, \quad p = 1, 2, \ldots, L_2 \tag{9}$$

And then the outputs of the second convolution layer are:

$$R_{i,h}^{l,p} = R_{i,h}^l * W_p^2 \tag{10}$$

After these two convolution layers, every $R_{i,h}$ has $L_1 L_2$ outputs. Each output is binarized by the Heaviside step function (one for positive input and zero otherwise) to obtain an integer value of each pixel of $R_{i,h}^l$, which is in the range $\left[0, 2^{L_2} - 1\right]$. Thus, we gain an integer-value image $T_{i,h}^l$

$$T_{i,h}^l = \sum_{p=1}^{L_2} 2^{p-1} H(R_{i,h}^l * W_p^2) \tag{11}$$

Further, $T_{i,h}^l$ is transformed into a histogram $\text{hist}\left(T_{i,h}^l\right)$. Then the feature of input $R_{i,h}$ is defined by PCANet as:

$$\kappa_{i,h} = \left[\text{hist}\left(T_{i,h}^1\right), \text{hist}\left(T_{i,h}^2\right), \ldots, \text{hist}\left(T_{i,h}^{L_1}\right)\right] \tag{12}$$

The features obtained as above are fed into a support vector machine (SVM) to train a model which can classify superpixels of intermediate class to CC or UC. It is worth noting that the CC of the first phase includes not only the changed pixels caused by real terrain variation, but also changed pixels caused by strong speckle noise.

*2.3. Second Phase Deep Learning*

As stated above, when SAR images are contaminated by strong speckle noise, the CC of the first phase contains two categories of change. One is false change caused by speckle noise called FCC, the other is caused by real terrain variation called RCC. Thus, in the second phase, we aim to separate FCC and RCC, between which the intra-class interval is so small that they are difficult to distinguish. However, the hypostatic difference between the two categories is such that the change caused by strong speckle noise has strong randomness. If the influence of the random noise can be greatly weakened, discrimination between the RCC and FCC can be increased. Therefore, in the second deep learning phase, we adopt different methods to the first phase. One key step in the second phase is speckle noise suppression based on low rank and sparse decomposition. Details are as follows.

2.3.1. Superpixel Generation on the Updated SAR Images

In the second phase, we firstly use mask processing on the original SAR images $I_1$ and $I_2$ to set the pixels classified as UC in the first phase to zero, thus, easing the burden on the classifier in this phase. Then SLIC is conducted on these two masked images to generate new superpixel objects denoted by $S_{b,i}^2$. The superpixel generation in the phase has two differences from that in the first phase. Firstly, the superpixel generation of this phase is based on the masked images, so the spatial context of the pixels has altered significantly leading to different superpixel patterns. Secondly, when applying SLIC in this phase, we set the number of pixels of each superpixel to be less than that in the first phase because there are many discontinuous areas caused by the mask operation compared to the generation in the first phase. Then we reshape the superpixel objects $S_{b,i}^2$ into vectors $V_{b,i,h}^2$ using a strategy similar to that in the first phase.

2.3.2. Low Rank and Sparse Decomposition

The principle of using LRSD is that the pair of noisy superpixels from the same unchanged area of $I_1$ and $I_2$, have an inherent large correlation with a low rank characteristic. Therefore, to discriminate RCC and FCC, we propose an idea based on LRSD to suppress speckle noise and restore the superpixel objects. The LRSD model establishes the effective expression of observed data with noise [33, 34]. Low rank regularization constraints and sparse regularization constraints can separate noise effectively from observed data and recover data. By optimizing the LRSD model, speckle noise can be separated and observed objects restored, which may greatly increase the discrimination between RCC and FCC.

At first, we apply a logarithmic operation on each vector of superpixel objects to convert multiplicative speckle noise to additive noise. Then, each vector can be formulated as follows.

$$V_{b,i,h}^2 = u_{b,i,h}^2 + e_{b,i,h}^2 \qquad (13)$$

Where $u_{b,i,h}^2$ indicates the scattering information of ground objects not polluted by speckle noise, and $e_{b,i,h}^2$ indicates additive speckle noise. All vectors $V_{1,i,h}^2$ and $V_{2,i,h}^2$ are arranged in pairs to construct a matrix $\boldsymbol{\Phi} = \left[ V_{1,1,1}^2, V_{2,1,1}^2, \ldots, V_{1,1,q_1}^2, V_{2,1,q_1}^2, \ldots\ldots, V_{1,v,1}^2, V_{2,v,1}^2, \ldots, V_{1,1,q_v}^2, V_{2,1,q_v}^2 \right]$, as shown in Fig. 5. Thus, we can obtain the matrix version of equation (13) as equation (14).

$$\boldsymbol{\Phi} = U + E \qquad (14)$$

Where, $U = \left[ u_{1,1,1}^2, u_{2,1,1}^2, \ldots, u_{1,1,q_1}^2, u_{2,1,q_1}^2, \ldots\ldots, u_{1,v,1}^2, u_{2,v,1}^2, \ldots, u_{1,1,q_v}^2, u_{2,1,q_v}^2 \right]$,
$E = \left[ e_{1,1,1}^2, e_{2,1,1}^2, \ldots, e_{1,1,q_1}^2, e_{2,1,q_1}^2, \ldots\ldots, e_{1,v,1}^2, e_{2,v,1}^2, \ldots, e_{1,1,q_v}^2, e_{2,1,q_v}^2 \right]$.

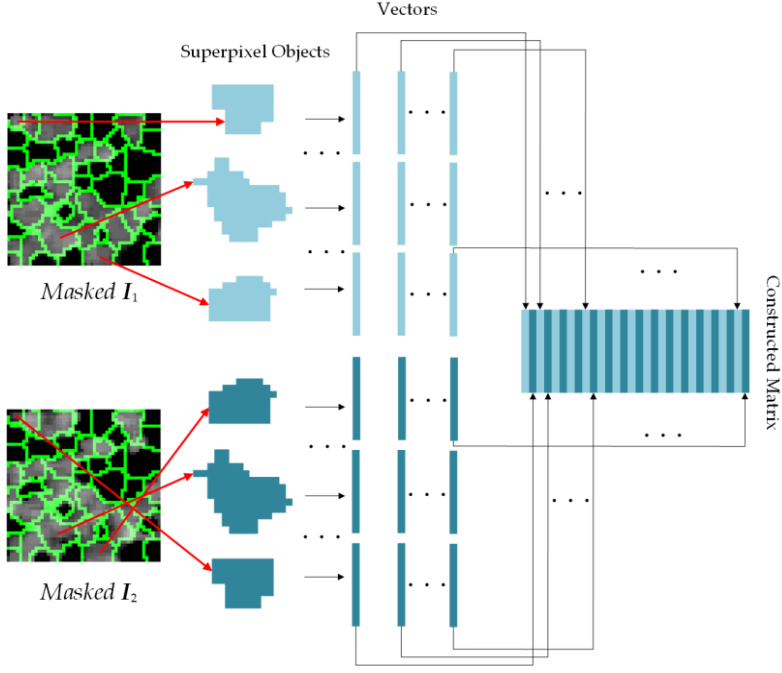

**Figure 5.** Construction of matrix $\boldsymbol{\Phi}$.

According to the principle of low rank representation, in order to estimate a low rank matrix $U$ and a spare matrix $E$ from a noise-contaminated observed $\boldsymbol{\Phi}$, we formulate an optimization problem as follows.

$$\min_{U,E} \|U\|_* + \varepsilon(1-\lambda)\|U\|_{2,1} + \varepsilon\lambda\|E\|_{2,1}, \quad \text{subject to } \boldsymbol{\Phi}=U+E \qquad (15)$$

Where $\|\cdot\|_*$ indicates the nuclear norm, $\|\cdot\|_{2,1}$ indicates the $l_1$ norm of a vector formed by the $l_2$ norm of the column vector of the underlying matrix. $\|\cdot\|_*$ induces sparsity of the singular values of the matrix, and $\|\cdot\|_{2,1}$ induces sparsity of the elements of the matrix.

The optimization problem can be solved by an augmented Lagrange algorithm. The augmented Lagrange formula of the problem (15) is as follows:

$$L(U,E,X,\mu) = \|U\|_* + \varepsilon(1-\lambda)\|U\|_{2,1} + \varepsilon\lambda\|E\|_{2,1} + \langle X, \boldsymbol{\Phi}\text{-}U\text{-}E\rangle + \frac{\mu}{2}\|\boldsymbol{\Phi}-U-E\|_F^2 \qquad (15)$$

Where $X$ is the Lagrange multiplier. Given $X = X_k$ and $\mu = \mu_k$, the key to solving the problem is to solve:

$$\min_{U,E} L(U,E,X_k;\mu_k) \qquad (16)$$

the solution of which will emerge though iteration. First, fix $U = U_k$, and solve:

$$E_{k+1} = \arg\min_{E} L(U_k, E, X_k; \mu_k) \qquad (17)$$

Then, fix $E = E_{k+1}$, and solve:

$$U_{k+1} = \arg\min_{U} L(U_k, E_{k+1}, X_k; \mu_k) \qquad (18)$$

After LRSD, we utilize column vectors $\boldsymbol{u}_{1,i,h}^2$ and $\boldsymbol{u}_{2,i,h}^2$ of low rank matrix $\boldsymbol{U}$ to restore $\boldsymbol{V}_{b,i,h}^2$, abandoning the noise matrix $\boldsymbol{E}$, as shown in Fig. 6.

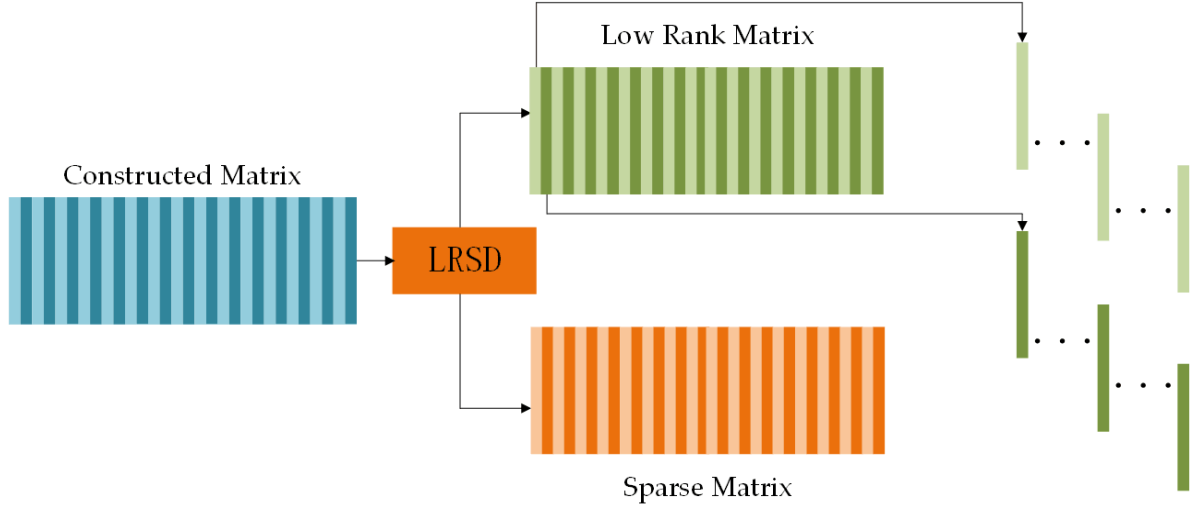

**Figure 6.** LRSD of the vectors from superpixel objects.

2.3.3. SPDI Generation and FCM

In the second phase, the difference vector is obtained from the superpixel vectors restored by LRSD, and FCM clustering is also adopted. At this stage, $\boldsymbol{F}_{i,h}^2 = \left|\boldsymbol{u}_{1,i,h}^2 - \boldsymbol{u}_{2,i,h}^2\right|$, forming a new SPDI, is taken as the input of FCM, to be clustered into three classes, FCC $\omega_{fc}^2$, RCC $\omega_{rc}^2$ and the intermediate class $\omega_{mc}^2$.

2.3.4. Training PCANet2 and Obtaining the Final Change Map

We use a PCANet to classify the vectors belonging to the intermediate class $\omega_{mc}^2$ to FCC or RCC. To discriminate it from PCANet1, we name it PCANet2, the structure of which is the same as PCANet1. Although they have the same network structure, their training data are very different, leading to distinct network parameters. Also, since the size of the superpixels of this phase is smaller than that in the first phase, the patch size of PCANet2 is smaller than that of PCANet1 relatively. The training of PCANet2 is similar to that of PCANet1. Once the network extracts the features of all the training samples, the extracted features are employed to train an SVM model, which classifies the intermediate class $\omega_{mc}^2$. In this way, we obtain the result of the second phase, which discriminates strong-noise-induced changes and real terrain changes. Finally, the real changed pixels of the SAR images are ideally only the pixels of superpixel objects belonging to RCC $\omega_{rc}^2$. By this, the final binary change detection result can be obtained.

**3. Experiments and Results**

To demonstrate the accuracy and effectiveness of the proposed approach, we compared TPOBDL with other state-of-the-art methods: principal component analysis and *k*-means clustering (PCAKM) [8], Gabor feature extraction and PCANet (GaborPCANet) [22], neighborhood-based ratio and extreme learning machine (NR_ELM) [35] and convolutional-wavelet neural network (CWNN)[36].

*3.1. Datasets and Experimental Setup*

We applied the proposed and benchmark methods to three real space-borne SAR datasets to evaluate the performance of TPOBDL. The three datasets used are co-registered and geometrically corrected SAR images acquired by the COSMO-Skymed satellite sensor, as shown in Fig. 7. The images in Fig. 7(a)(b)(c) were acquired on June 10, 2016 and those in Fig. 7(d)(e)(f) on April 26, 2017. The three areas are selected to represent different landscapes containing a river, a plain, mountain and buildings. They are all of size 400 × 400 pixels. It is obvious that the three SAR datasets suffer from speckle noise. Many speckle noise reduction filters exist, but the balance between speckle suppression and detail preservation remains challenging [35]. Therefore, no speckle filters were applied to the three SAR datasets. The corresponding ground reference datasets are shown in Fig. 7(g)(h)(i).

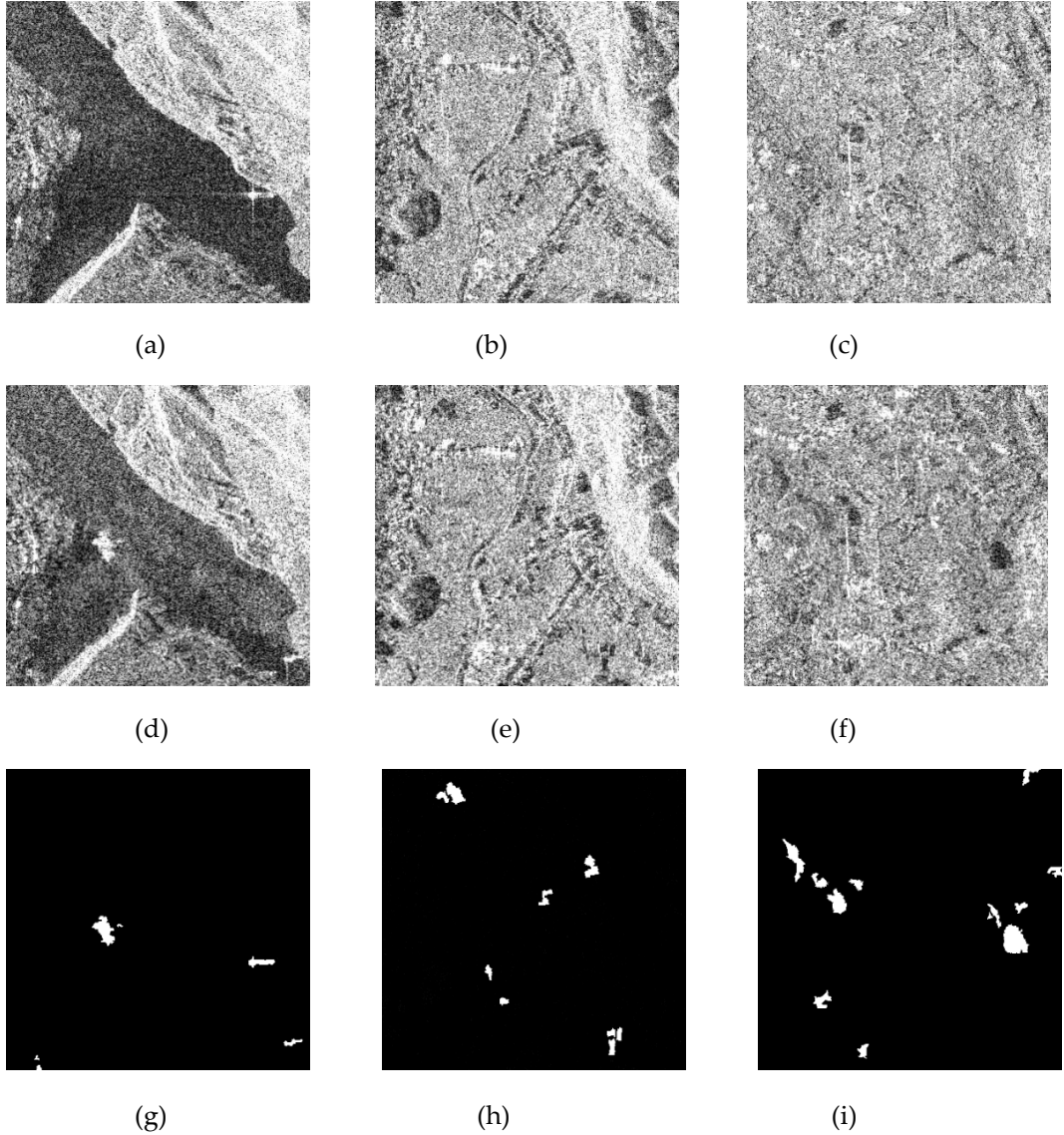

**Figure 7.** (a)(d) are dataset C1 that contains river and mountains, and (g) is its ground truth. (b)(e) are dataset C2 that contains buildings, roads and mountains , and (h) is its ground truth. (c)(f) are dataset C3 that contains plain and buildings, and (i) is its ground truth.

How to evaluate the performance of SAR image change detection algorithms is a key issue. Here, we utilized several state-of-the-art evaluation metrics, including the false alarm probability $P_f$ ,

missing detection probability $P_m$, percentage correct classification $PCC$, Kappa coefficient $KC$ and $GD/OE$ [1,22]. Assume that the actual numbers of pixels belonging to UC and CC are denoted by $N_u$ and $N_c$, respectively, in the ground reference data, then

$$P_f = \frac{F_n}{N_u} \times 100\% \tag{19}$$

$$P_m = \frac{M_n}{N_c} \times 100\% \tag{20}$$

Where $F_n$ denotes the number of unchanged pixels detected as changed, while $M_n$ represents the number of changed pixels detected as unchanged.

$$PCC = \frac{(N_u + N_c - F_n - M_n)}{N_u + N_c} \times 100\% \tag{21}$$

$$KC = \frac{(PCC - PRE)}{1 - PRE} \times 100\% \tag{22}$$

where,

$$PRE = \frac{(N_c + F_n - M_n) \times N_c + (N_u + M - F) \times N_u}{(N_c + N_u)^2} \tag{23}$$

The definition of $GD/OE$ is then as follows.

$$GD/OE = \frac{(N_u - M_n)}{F_n + M_n} \times 100\% \tag{24}$$

*3.2. Experiments*

We analyzed and evaluated the final results visually and quantitatively.

The change detection results of multi-temporal SAR dataset C1 are shown in Fig. 8 and Table 1. As presented in Fig. 8, the change map of PCAKM contains many false alarms, scattered widely across the image with $P_f$ reaching 39.23%. This is because PCAKM is unable to classify the false changes caused by strong speckle noise and real changes caused by terrain variation as shown in Fig. 8 (a). However, different from PCAKM, the false alarms of GaborPCANet, NR_ELM and CWNN are centred in the river, as shown in Fig.8 (b)(c)(d). On one hand, PCAKM uses *pixel* values for change detection, which are affected by strong speckle noise. Thus, the $P_f$ of PCAKM is very high. However, GaborPCANet and CWNN, two deep learning-based methods, can extract deep features and have a certain speckle noise suppression capability, so the $P_f$ is greatly reduced compared to PCAKM. Moreover, the extreme learning machine in NR_ELM can also effectively extract features and suppress speckle noise. Therefore, the performance of GaborPCANet, NR_ELM and CWNN is better than that of PCAKM. On the other hand, compared to the original two SAR images, we found that false alarms occur in the river region for the latter three methods. The river region in the two SAR images looks very dark, because the river backscatter of electromagnetic waves is relatively weak. Thus, under strong speckle noise, the signal-to-noise ratio (SNR) in the river region of the SAR image is very low. Therefore, in this case, the difference in values of pixels between the two images in the river region is relatively large, and pixels in the river region are easily classified as CC.

It can be seen that the final change map obtained by the proposed approach TPOBDL is very close to the ground reference, as shown in Fig. 8 (f). Compared with the former methods, the $P_f$ obtained by TPOBDL is only 0.18% (see Table 1), which is a remarkable result. This is because the second phase of TPOBDL uses a special network to identify the pixels of FCC and those of RCC. In addition, compared to CWNN, our approach uses object-based deep learning removing those scattered false alarms effectively, which demonstrates the advantages of object-based deep learning. Therefore, TPOBDL can eliminate effectively the false alarms caused by strong speckle noise.

As can be seen from Table 1, the quantitative analysis is consistent with the visual analysis. The performance of TPOBDL is better than for the benchmark algorithms in terms of $PCC$, $P_f$, $KC$ and $GD/OE$. It is worth noting that although the $P_m$ of PCAKM, GaborPCANet and NR_ELM are smaller than that of TPOBDL, these three methods come at the cost of a much larger $P_f$. The reason why the $P_m$ of our method is larger than for the three benchmark methods, is that a few superpixel objects of RCC are mistakenly classified as FCC in the second deep learning phase. Therefore, we need to consider the value of the more convincing $KC$. TPOBDL has the highest value of $KC$ (97.84%), which means that the change detection accuracy of TPOBDL is the highest amongst all five methods.

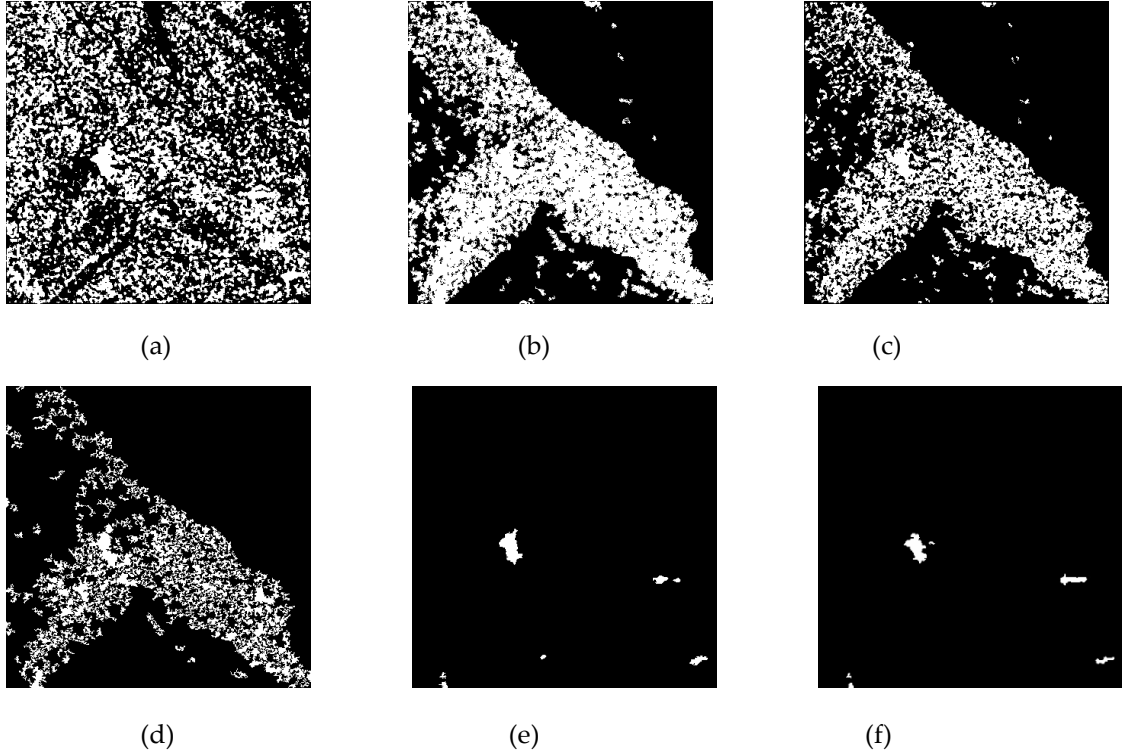

**Figure 8.** Results of experiments on C1; (a) PCAKM; (b) GaborPCANet; (c) NR_ELM; (d) CWNN; (e) TPOBDL; (f) ground truth.

**Table 1.** Comparison of evaluation metrics amongst PCAKM, GaborPCANet, NR_ELM, CWNN and TPOBDL on dataset C1 using the false alarm probability ($P_f$), missing detection probability ($P_m$), percentage correct classification ($PCC$), Kappa coefficient ($KC$) and $GD/OE$.

| Methods | Results on C1(%) | | | | |
|---|---|---|---|---|---|
| | PCC | $P_f$ | $P_m$ | GD/OE | KC |
| PCAKM[9] | 60.99 | 39.24 | 1.78 | 0.07 | 58.87 |
| GaborPCANet[23] | 64.67 | 35.46 | 4.88 | 0.08 | 59.36 |
| NR_ELM[33] | 73.85 | 26.26 | 9.86 | 0.11 | 61.39 |
| CWNN[34] | 85.22 | 14.69 | 29.18 | 0.19 | 65.67 |
| TPOBDL | 99.71 | 0.18 | 15.10 | 9.97 | 97.84 |

Fig. 9 and Table 2 present the final change detection results on dataset C2. In terms of visual comparison, PCAKM still includes many false alarms. The performance of GaborPCANet is better than that of PCAKM in terms of $P_f$. However, there are several false alarms due to speckle noise. Moreover, for each of PCAKM, GaborPCANet or NR_ELM, there is an obvious long and narrow area with fewer false alarms in the upper right corner of the change map. Comparing the original two multi-temporal SAR images, we find that this long and narrow area has an area of relatively strong back-scattering (visually white), which means the amplitude value of these pixels is relatively large. This indicates that change detection in areas with strong scattering is less affected by speckle noise because of the high SNR. This situation is exactly the opposite of the high false alarm phenomenon in the river region in the experiments on C1. As for CWNN, it is clear that the value of $P_f$ due to speckle noise is smaller than for the three benchmarks. This benefit arises from the wavelet pooling layers in CWNN, which suppress speckle noise by losing high-frequency sub-bands while preserving low-frequency sub-bands to extract features. However, TPOBDL has less false alarms than CWNN, because the object-based methodology is adopted, which greatly reduces classification uncertainty induced by rectangular patches. As for TPOBDL, two-phase deep learning is not only effective for change detection in low SNR region, but also for change detection in high SNR regions. This is due to the influence of the LRSD, which greatly constrains the influence of speckle noise. Among the five methods, TPOBDL has the best performance in terms of $PCC$, $P_f$, $GD/OE$ and $KC$, reaching 99.43%, 0.26%, 4.70% and 95.67%, respectively.

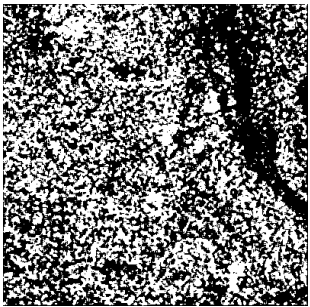 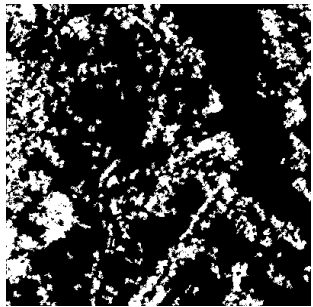 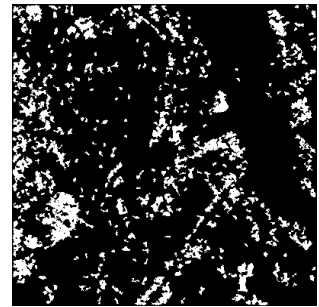

(a)          (b)          (c)

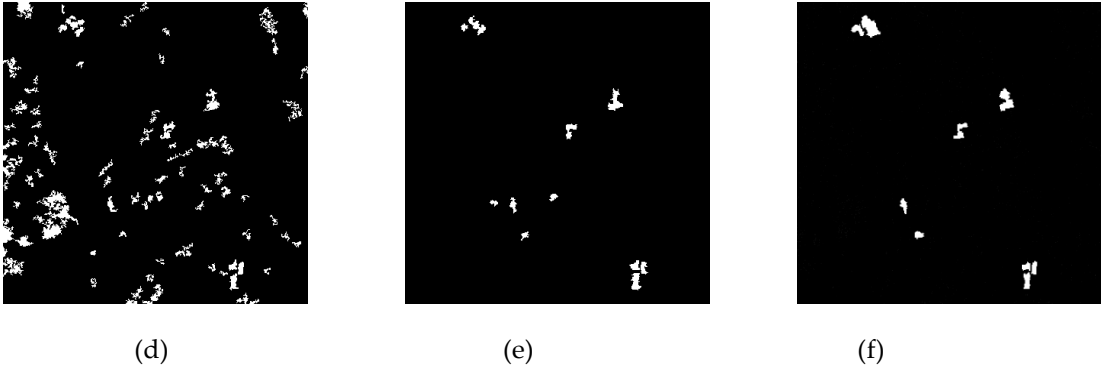

(d)                  (e)                  (f)

**Figure 9.** Results of experiments on C2; (a) PCAKM; (b) PCANet; (c) NR_ELM;(d) CWNN; (e) TPOBDL; (f) ground truth.

**Table 2.** Comparison of evaluation metrics amongst PCAKM, GaborPCANet, NR_ELM, CWNN and TPOBDL on dataset C2 using the false alarm probability ($P_f$), missing detection probability ($P_m$), percentage correct classification ($PCC$), Kappa coefficient ($KC$) and $GD/OE$.

| Methods | Results on C2(%) | | | | |
|---|---|---|---|---|---|
| | $PCC$ | $P_f$ | $P_m$ | $GD/OE$ | $KC$ |
| PCAKM[9] | 55.65 | 45.24 | 1.81 | 0.07 | 58.13 |
| GaborPCANet[23] | 79.64 | 20.66 | 6.19 | 0.14 | 63.22 |
| NR_ELM[33] | 86.99 | 13.14 | 7.11 | 0.21 | 67.37 |
| CWNN[34] | 95.24 | 4.59 | 12.41 | 0.56 | 78.49 |
| TPOBDL | 99.43 | 0.26 | 15.02 | 4.70 | 95.67 |

The results of experiments on dataset C3 are exhibited in Fig. 10 and Table 3. The performance of PCAKM is again the least good. Compared with the first two datasets, there are no weak backscattering regions (like river, C1) or strong backscattering regions (like mountain, C2). However, the contrast in the whole scene of C3 is relatively low, which means that classification may be more challenging due to low discrimination. Thus, it can be seen from Table 3 that the $P_m$ of all methods is relatively high. Still, TPOBDL is superior to CWNN in terms of $P_m$ under the circumstances, which is opposite to the experiments on C1 and C2. Among the five methods, TPOBDL again produces the best result, with a $PCC$ of 98.42%, $P_f$ of 1.18%, $GD/OE$ of 1.59% and $KC$ of 89.32%. It is worth noting that in the experiments on C3, TPOBDL again produces the best values of $PCC$, $P_f$ and $KC$, while also producing a similar $P_m$ of 19.64% to other methods, at the same time. The experimental results illustrate the superiority of TPOBDL

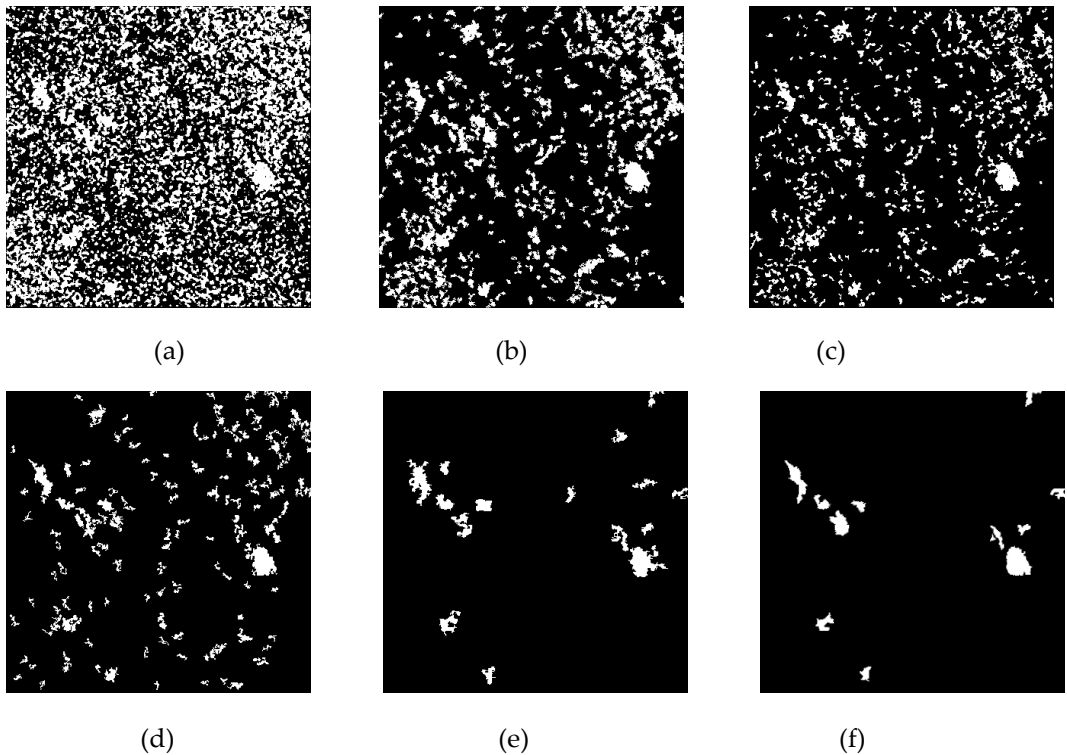

(a)                      (b)                      (c)

(d)                      (e)                      (f)

**Figure 10.** Results of experiments on C3; (a) PCAKM; (b) PCANet; (c) NR_ELM; (d) CWNN; (e) TPOBDL; (f) ground truth.

**Table 3.** Comparison of evaluation metrics amongst PCAKM, GaborPCANet, NR_ELM, CWNN and TPOBDL on dataset C3 using the false alarm probability ($P_f$), missing detection probability ($P_m$), percentage correct classification ($PCC$), Kappa coefficient ($KC$) and $GD/OE$.

| Methods | Results on C3(%) | | | | |
| --- | --- | --- | --- | --- | --- |
|  | $PCC$ | $P_f$ | $P_m$ | $GD/OE$ | $KC$ |
| PCAKM[9] | 62.23 | 38.29 | 14.39 | 0.07 | 58.50 |
| GaborPCANet[23] | 84.61 | 15.32 | 18.92 | 0.16 | 64.84 |
| NR_ELM[33] | 89.54 | 9.98 | 31.90 | 0.21 | 67.56 |
| CWNN[34] | 94.53 | 5.02 | 25.90 | 0.43 | 75.55 |
| TPOBDL | 98.42 | 1.18 | 19.64 | 1.59 | 89.32 |

## 4. Discussion

### 4.1. Parameters Selection

In the proposed approach, there exist four parameters to be discussed, which are the number of superpixels $SP_1$ and the patch size $k_1$ in the first phase, and the equivalents, $SP_2$ and $k_2$, in the second phase. These four parameters affect the ability to learn neighborhood information in the two-phase object-based deep learning approach. As indicated in [21], when the patch size is set as $5 \times 5$

, it leads to an optimal result. Hence, we fix $k_1=5$ at the beginning. As for $SP_1$ and $SP_2$, to reduce redundancy and increase superpixel generation efficiency, we assume $SP_i \approx (M \times N)/k_i^2$ $(i=1,2)$, which means that the number of pixels in a superpixel and the number of pixels in a patch should be the same, as far as possible. So we fix $SP_1=6400$. Then, we conduct experiments on $SP_2=$ 17800, 6400, 3200 and $k_2=3,5,7,9$ in pair-wise fashion, respectively. The experimental results are shown in Fig. 11-12.

Observing from Fig. 11-12, we found that when $SP_2=17800$ and $k_2=3$, the values of $PCC$ and $KC$ were the best. The experimental result is consistent with the principle of the proposed approach. As mentioned before, the spatial context of the pixels has altered significantly after masking in the second phase. And, there may be many discontinuous areas after masking. Hence, superpixel objects with a small number of pixels have the benefit of avoiding heterogeneous pixels inside the objects, which reduces classification uncertainty in PCANet2. This reveals that, in the second phase, the relatively small superpixels helps the PCANet2 to exploit more details, which cater to the purpose of distinguishing RCC and FCC.

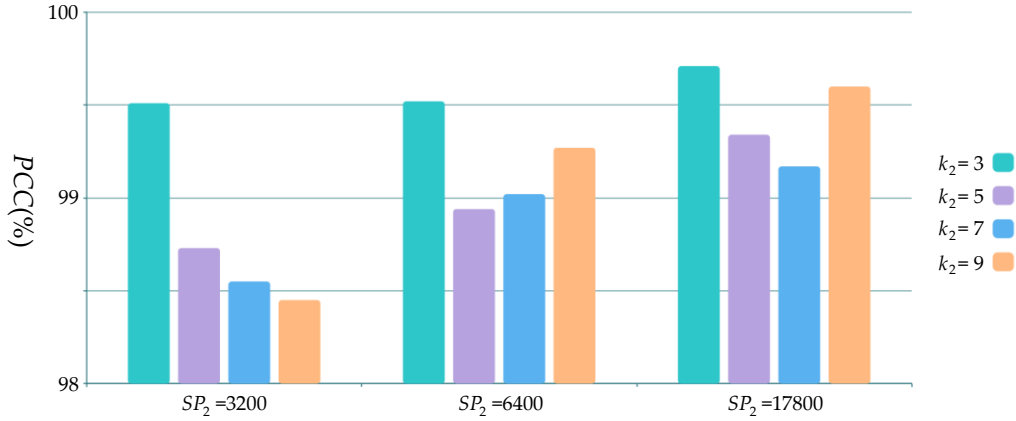

**Figure 11.** The influence of different parameters ($SP_2$ and $k_2$) on PCC.

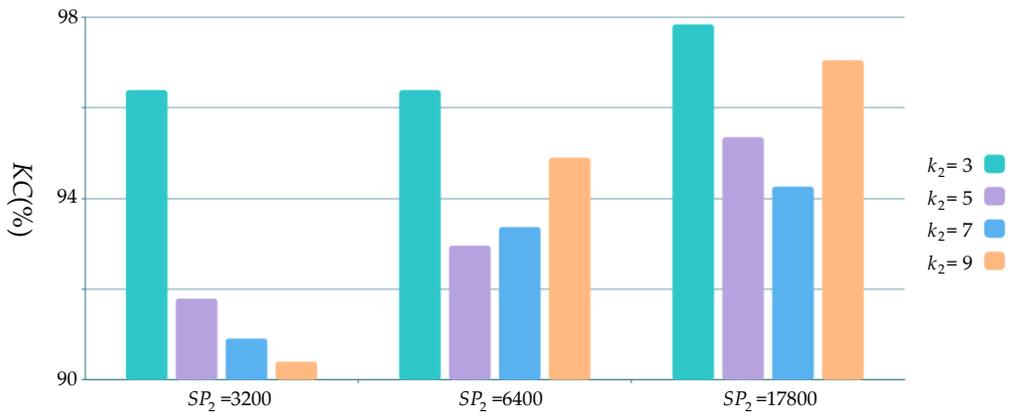

**Figure 12.** The influence of different parameters ($SP_2$ and $k_2$) on KC.

We then fixed the parameters of the second phase as $SP_2=17800$ and $k_2=3$ to conduct experiments on $SP_1=17800, 6400, 3200$ and $k_1=3,5,7,9$ in a pair-wise fashion, respectively. The experimental results are presented in Fig. 13-14.

As shown in Fig. 13-14, there are two pairs of $SP_1$ and $k_1$ that obtain a larger PCC and KC than other parameter values. One pair is $SP_1 = 6400$ and $k_1 = 5$, and the other pair is $SP_1 = 3200$ and $k_1 = 7$. This means that superpixels with relatively large number of pixels are of benefit for classifying UC and CC in the first phase. After further observation, these two pairs of parameters adhere to $SP_i \approx (M \times N)/k_i^2$, which indicates that theoretically the number of pixels in a superpixel should be similar to the number of pixels in a patch. Thus, the best parameter combination is $SP_1 = 3200$, $k_1 = 7$ for the first phase, and $SP_2 = 17800$, $k_2 = 3$ for the second phase.

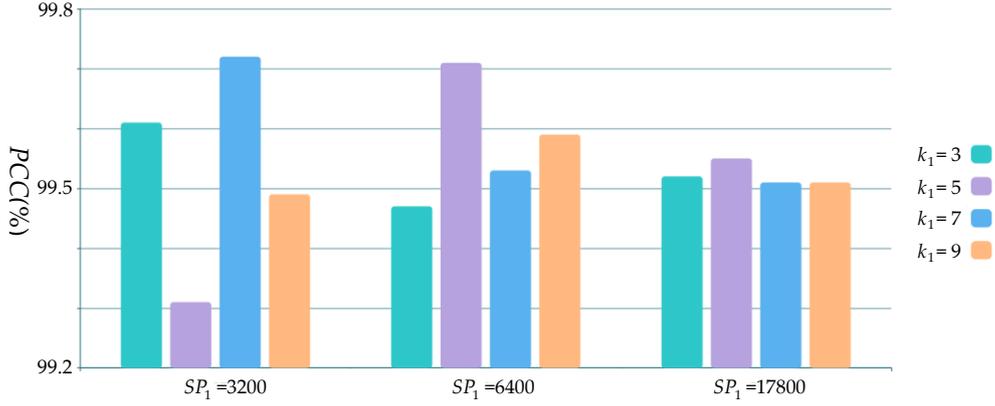

**Figure 13.** The influence of different parameters ($SP_1$ and $k_1$) on PCC.

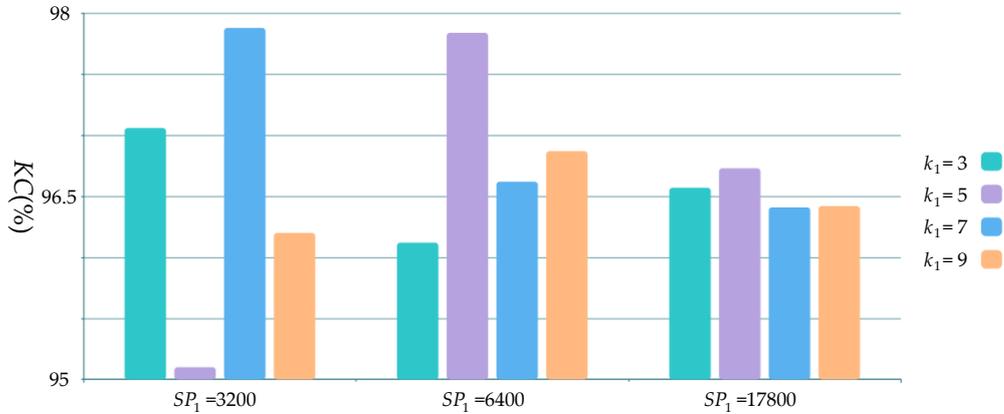

**Figure 14.** The influence of different parameters ($SP_1$ and $k_1$) on KC.

*4.2. Comparison with Other Methods*

Firstly, we compare the proposed approach with four other methods. The experimental results of all methods are presented in Fig. 8-10 and Tables 1-3. TPOBDL outperforms other methods in all evaluation indicators, except for missing alarms rate. This is because by using superpixel objects and two phases of PCANet, TPOBDL is more robust to speckle noise, able to extract deep features and capable of learning the nonlinear relations from multi-temporal SAR images efficiently. The patches reshaped from superpixel objects with homogeneous pixels are beneficial to the deep feature extraction and PCANet training, which avoids uncertainty due to rectangular patches.

The two deep learning phases in TPOBDL are important for acquiring the desired change detection performance. The first phase generally classifies pixels into two classes, CC and UC. However, there are actually two kinds of changes in CC. One is strong speckle noise-induced change, and the other is real terrain variation-induced change. In the second phase, the pixels belonging to UC are set to zero so that the PCANet2 can focus on identifying two indistinguishable changes. PCANet2 faces a more difficult classification tasks than PCANet1. Hence, we equip the second phase with LRSD to suppress noise and increase the ability to discriminate the two previously indistinguishable changes. Despite noise interference, multi-temporal SAR images of the same object should have a strong correlation. Based on this principle, we established the LRSD model. LRSD can not only suppress speckle noise, but also highlight the correlation between objects via the low rank constraint, as shown in Fig. 15. Through this, TPOBDL achieves the best performance amongst the five methods when facing strong speckle noise. It is worth noting that there is no speckle filtering in TPOBDL.

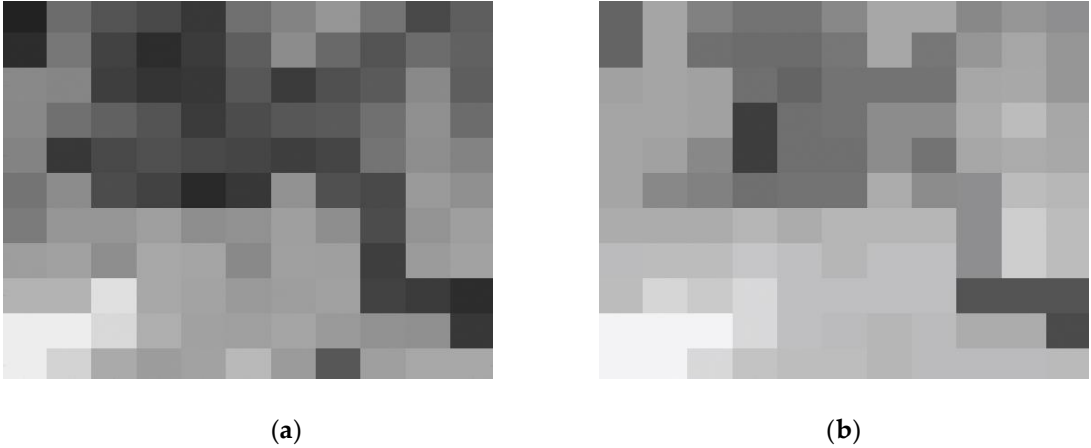

(**a**)  (**b**)

**Figure 15.** (a) A selected object before LRSD; (b) The object after LRSD.

*4.3 Modular Deep Learning Framework for change detection*

In the proposed approach, PCANet1 in the first phase completes the classification tasks of CC and UC, and PCANet2 in the second phase completes the classification tasks of RCC and FCC. In fact, other deep neural networks can also be used in the first stage, instead of PCANet. In the same way, it is not necessary to use the PCANet in the second phase. Therefore, the two phase deep learning framework proposed in this paper can be regarded as a modular structure. The structure does not actually limit what deep learning models are used. The key to this modular structure is hierarchical classification. Moreover, the advantage of this modular deep learning framework is that the deep neural network in each module can complete a specialized, and not particularly complicated task, so the difficulty of classification in each module is reduced. For example, in this research, if only one PCANet is used to complete the classification of UC, RCC and FCC simultaneously, it is easy to generate more misclassifications, which will lead to a larger number of false alarms or larger number of missing alarms. In addition, this modular deep learning-based change detection structure is particularly suitable for engineering implementation.

*4.4 Time- series SAR Images to Suppress Speckle Noise*

As is commonly known, speckle noise is a major factor affecting the detection of changes in multi-temporal SAR images. Although speckle filtering is not used in the proposed approach, it is conceivable that the performance of the algorithm will be improved to a certain extent if speckle noise filtering is performed before change detection. Since multi-temporal SAR images can be regarded as a time-series, and the images of the same observation scene are coherent, speckle noise can be suppressed based on this principle effectively. How to suppress speckle noise using time-series SAR images is also a topic for subsequent research.

## 5. Conclusions

In this research, a novel change detection algorithm with two-phase object-based deep learning approach for multi-temporal SAR images is presented. An object-based approach is used instead of a pixel-wise approach. The object-based change detection approach can effectively exploit the spatial context of neighborhood pixels, which is conducive to increasing the ability to identify UC and CC. Using superpixel objects, the pixels in each object are generally more homogeneous, which avoids the classification uncertainty caused by heterogeneous pixels and provides high-quality training samples for subsequent PCANets. In addition, this paper uses a two-phase deep learning framework to implement change detection on multi-temporal SAR images. The first phase of deep learning realizes the distinction between UC and CC. The second phase of deep learning realizes the distinction between RCC and FCC. The two-phase deep learning framework can tackle effectively the classification challenge faced by deep learning in each phase, and can effectively distinguish RCC and FCC, while maintaining a very low false alarm under strong speckle noise. The experimental results illustrate that the proposed approach can achieve high accuracy and validity.

**Acknowledgements**

This research was funded by the National Science Foundation of China under Grants No.61301224. This research was also partly supported by the Basic and Advanced Research Project in Chongqing under Grants No. cstc2017jcyjA1378 and No. cstc2016jcyjA0457.

**Author Contribution**

Xinzheng Zhang and Guo Liu conceived and designed the shceme. Guo Liu conducted experiments. Xinzheng Zhang, Peter M Atkinson and Ce Zhang analysed and discussed the results. Xinzheng Zhang and Guo Liu wrote the first draft. Xinzheng Zhang, Ce Zhang and Peter M Atkinson completed the revised paper. Xiaoheng Tan ,Xin Jian, Xichuan Zhou and Yongming Li gave some suggestions for the paper.